\documentclass[journal]{IEEEtran}
\usepackage{amsmath}
\usepackage{float}
\usepackage{graphicx}
\usepackage{subcaption}
\usepackage{textcomp}
\usepackage{xcolor}
\usepackage{float}
\usepackage{multirow}
\usepackage{csquotes}
\usepackage{url}
\usepackage{tabularx}
\usepackage{booktabs}
\newcolumntype{Y}{>{\raggedleft\let\newline\\\arraybackslash\hspace{0pt}}X}
\newcolumntype{Z}{>{\centering\let\newline\\\arraybackslash\hspace{0pt}}X}

\usepackage{enumitem}
\setlist[itemize]{noitemsep,topsep=0pt}

\ifCLASSINFOpdf

\else

\fi

\begin{document}

\title{Synthetic Periocular Iris PAI from a Small Set of Near-Infrared-Images.}

\author{Jose Maureira,
        Juan Tapia,~\IEEEmembership{Member,IEEE,}
        Claudia Arellano,~\IEEEmembership{Member,IEEE,}
        and~Christoph Busch,~\IEEEmembership{Senior Member,~IEEE}
\thanks{Juan Tapia and Christoph Busch, da/sec-Biometrics and Internet Security Research Group, Hochschule Darmstadt, Germany}.
\thanks{Jose Maureira is with Universidad de Santiago-Chile. Claudia Arellano is with Universidad Adolfo Ibañez, Santiago-Chile.}
\thanks{Corresponding Author: Juan Tapia. email: juan.tapia-farias@h-da.de}\\

\textbf{This work has been submitted to the IEEE for possible publication. Copyright may be transferred without notice, after which this version may no longer be accessible}

\thanks{Manuscript received July xx, 2021; revised xxxx.}}

\markboth{Journal of \LaTeX\ Class Files,~Vol.~14, No.~8, August~2015}%
{Shell \MakeLowercase{\textit{et al.}}: Bare Demo of IEEEtran.cls for IEEE Journals}

\maketitle

\begin{abstract}
Biometric has been increasing in relevance these days since it can be used for several applications such as access control for instance. Unfortunately, with the increased deployment of biometric applications we observe an increase of attacks. Therefore, algorithms to detect such attacks (Presentention Attack Detection (PAD)) have been increasing in relevance. The LivDet-2020 competition which focus on Presentation Attacks Detection (PAD) algorithms have shown still open problems, specially for unknown attacks scenarios.  In order to improve robustness of biometric systems, it is crucial to improve PAD methods. This  can be achieved by augmenting the number of presentation attack instruments (PAI) and bona fide images that are used to train such algorithms. Unfortunately, the capture and creation of presentation attack instruments and even the capture of bona fide images  is sometimes complex to achieve. This paper propose a novel PAI synthetically created (SPI-PAI) using four state of the art GAN algorithms (cGAN, WGAN, WGAN-GP and StyleGAN2) and a small set of periocular NIR images. A benchmark between GAN algorithms is performed using the Frechet Inception Distance (FID) between the generated images and the original images used for training. The best PAD algorithm reported by the LivDet-2020 competition was tested for us using the synthetic PAI which was obtained with the StyleGAN2 algorithm. Surprisingly, The PAD algorithm was not able to detect the synthetic images as a Presentation Attack, categorizing all of them as bona fide. Such results demonstrated the feasibility of synthetic images to fool presentation attacks detection algorithms and the need for such algorithms to be constantly updated and trained with larger number of images and PAI scenarios. 

\end{abstract}

\begin{IEEEkeywords}
Biometrics, Iris, Presentation Attack Detection, Presentation Attack Instrument, Synthetics images. 
\end{IEEEkeywords}

\IEEEpeerreviewmaketitle

\section{Introduction}
\label{sec:introduction}

\IEEEPARstart{I}ris recognition systems has been increasing in robustness over time. They are also affordable, non-invasive, and touchless; these strengths have promoted its popularity in the market in the recent years. Iris recognition systems are usually based on eye images captured using near-infrared (NIR) lighting and sensors. Although the constant development of new algorithms and their increasing improvement they are still susceptible to Presentation Attack Instruments (PAI)~\cite{OpenSource}. Where PAI corresponds to a biometric characteristic or object used in a presentation attack. In other words, a Presentation attack instrument is any set of characteristic or object, such images for instance, that are used to attack and fool a biometric system. 

There are several techniques to create Presentation Attack Images. Printed images, for instance, are easy to reproduce with different kinds of paper. PAIs can be also created using contact lenses, cosmetic lenses or plastic lenses which are easily available from different brands, although harder to get than just printing images\cite{bsif_bowyer}.

Presentation Attack Detection (PAD), on the other hand, refers to the ability of a biometric system to recognize PAIs, that would otherwise fool the system into recognizing an illegitimate user as a genuine one, by means of presenting a synthetic forged version of the original biometric trait to the capture device. The biometric community, including researchers and vendors, have thrown themselves into the challenging task of proposing and developing efficient protection mechanisms against the threat that PAIs represents~\cite{Galbally}. Attacks to biometric systems are not restricted to merely theoretical or academic scenarios anymore, as they are starting to be carried out against real-life operations. One example is the hacking of Samsung Galaxy S8 devices with the iris unlock system, using a regular printer and a contact lens. This case has been reported to the public from hacking groups attempting to get recognition for real criminal cases, including from live biometric demonstrations at conferences \footnote{\url{https://www.forbes.com/sites/ianmorris/2017/05/23/samsung-galaxy-s8-iris-scanner-hacked-in-three-simple-steps/#33f150b2ccba}}.

Results from the LivDet-2020~\cite{livdet2020} competition indicate that the development of iris PAD is still far from a fully solved research problem. There is a large difference in accuracy among baseline algorithms since they are mostly trained with significantly different data. Therefore, it is important to  have access to larger and diversified training datasets. Such datasets need to include larger number of bona fide images and PAIs to be able to train more robust PAD systems, considering known and unknown attacks.

This paper proposes the creation of synthetic periocular NIR iris images that can be used as PAI. Four state-of-the-art GAN's methods are used to synthesize such images. The FID metric is used to compare the resulting data set with the original mages from where those images are synthesized. In order to test if the resulting images can fool a PAD system, the winner algorithm from the LivDet-2020 competition is used \cite{livdet2020}. 

The rest of the article is organized as follows: Section~\ref{sec:relate} summarizes the related works on Presentation Attack Detection, PAI creation and Generative Adversarial Networks. The work proposed in this paper is presented in Section ~\ref{sec:proposal}. Section~\ref{sec:exp_results} describes the experiments and results obtained while Section~\ref{sec:conclusion} and Section~\ref{sec:future} present the conclusions and future work respectively.

\section{Related Work}
\label{sec:relate}
\subsection{Presentation Attack Detection (PAD)}

There are a vast number of research regarding to PAD algorithms. Hu et al.~\cite{hu2016iris}, proposed a Regional PAD, where regional features are extracted from local neighborhoods. This method is based on spatial pyramid (multi-level resolution) and relational measures (convolution on features with variable-size kernels). Several feature extractors, such as Local Binary Patterns (LBP), Local Phase Quantization (LPQ), and intensity correlogram are examined. The best performance is obtained using a three-scale LBP-based feature. Nguyen et al.~\cite{PNguyen} also proposed a PAD method by combining features extracted from local and global iris regions. First, they trained multiple VGG19~\cite{simonyan2015deep} networks from scratch for different iris regions. Then, the features were separately extracted from the last fully connected layer, before the classification layer of the trained models. The experimental results showed that the PAD performance was improved by fusing the features based on both feature-level and score-level fusion rules.

Gragnaniello et al.~\cite{gragnaniello2016using} have explored liveness detection in order to recognize attack presentation. They have found that the sclera region contains important information about iris liveness (SIDPAD). Hence, the authors extracted features from both the iris and sclera regions. The two regions are first segmented, and scale-invariant local descriptors (SID) are applied. A bag-of-feature method was then used to summarize the features. A linear Support Vector Machine (SVM) was used to perform final prediction. In~\cite{gragnaniello2016biometric}, the authors use domain-specific knowledge of iris PAD to incorporate it into the design of their prediction model (DACNN). With the domain knowledge, a compact network architecture is obtained, and regularization terms are added to the loss function to enforce high-pass/low-pass behavior. The authors demonstrated that this method can detect both face and iris attacks.

Yadav et al.~\cite{Yadav_2018_CVPR_Workshops}, reported a novel method where a combination of handcrafted and deep-learning-based features were used for iris PAD. They have fused multi-level Haralick features with VGG16 features to encode the iris textural patterns. The VGG16 features were extracted from the last fully connected layer, with a size of 4,096, and then reduced to dimensional vector by Principal Component Analysis (PCA). 

\begin{figure*}[!htb]
\centering
    \subcaptionbox{\centering Bona fide / Real}{\includegraphics[width=0.24\textwidth]{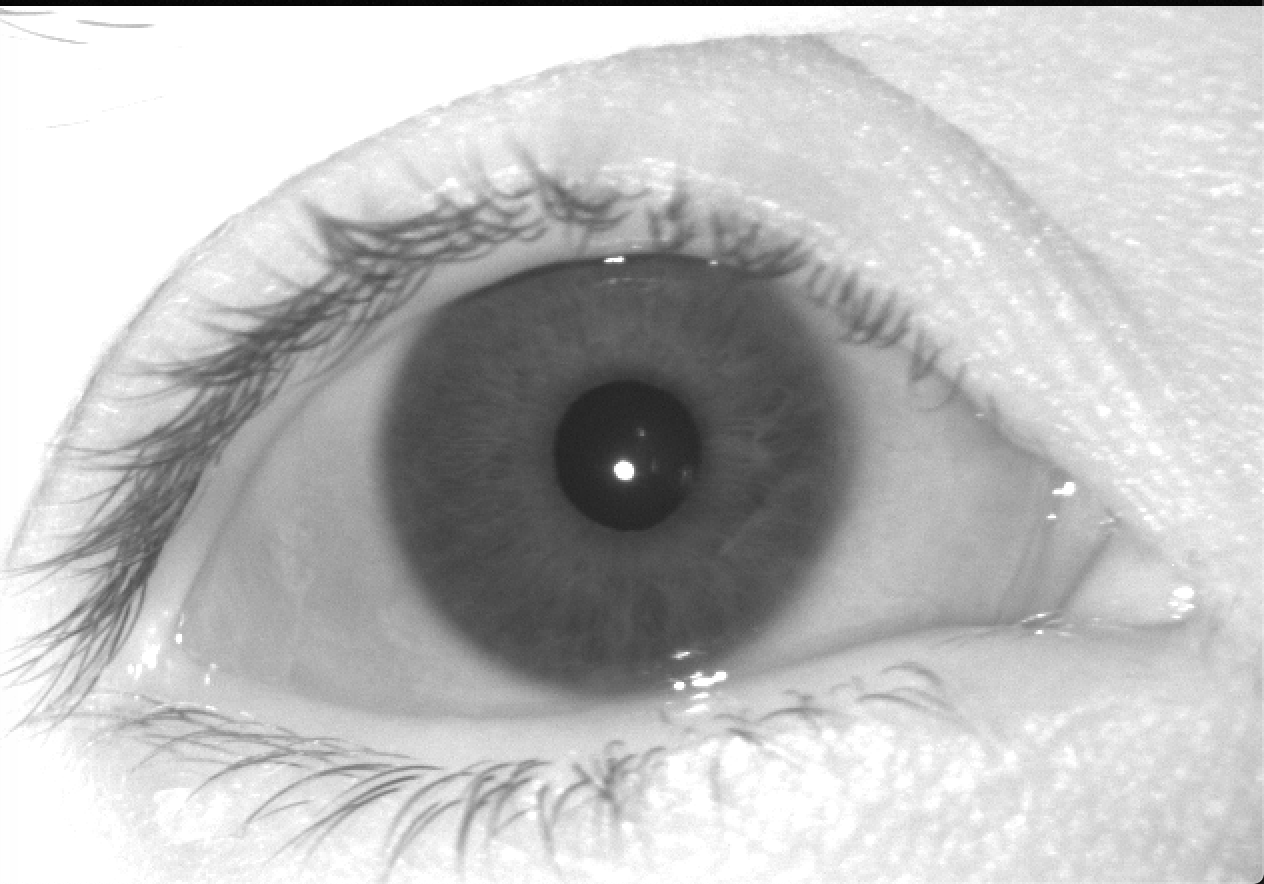}}
    \hfill
    \subcaptionbox{\centering Print-out LivDet-2020-Iris 2020}{\includegraphics[width=0.24\textwidth]{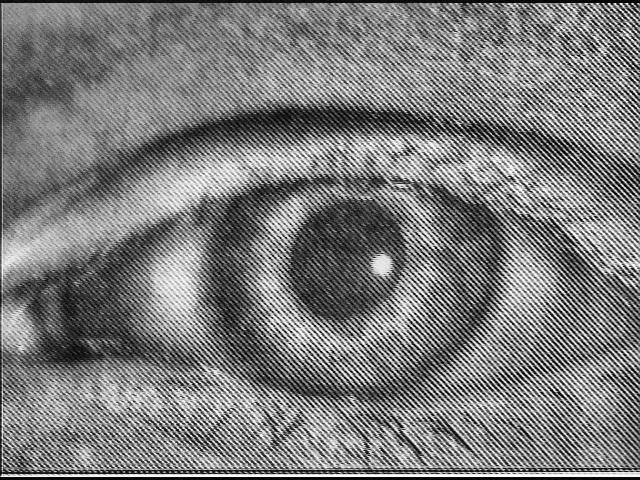}}
    \hfill 
    \subcaptionbox{\centering Cadaver (post-mortem subject) eye}{\includegraphics[width=0.24\textwidth]{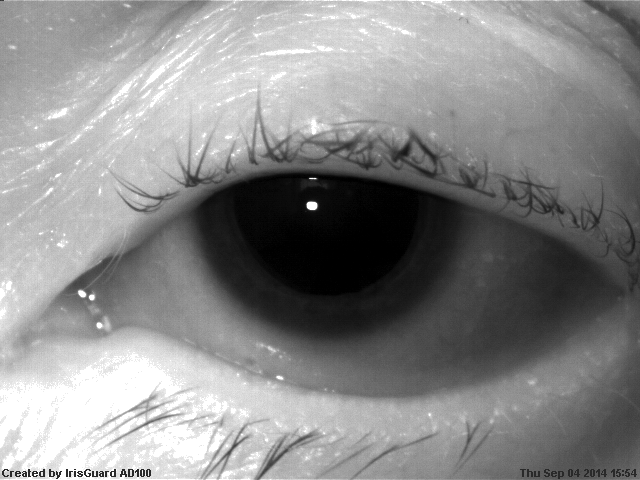}}
    \hfill
    \subcaptionbox{\centering Cosmetic contact lens}{\includegraphics[width=0.24\textwidth]{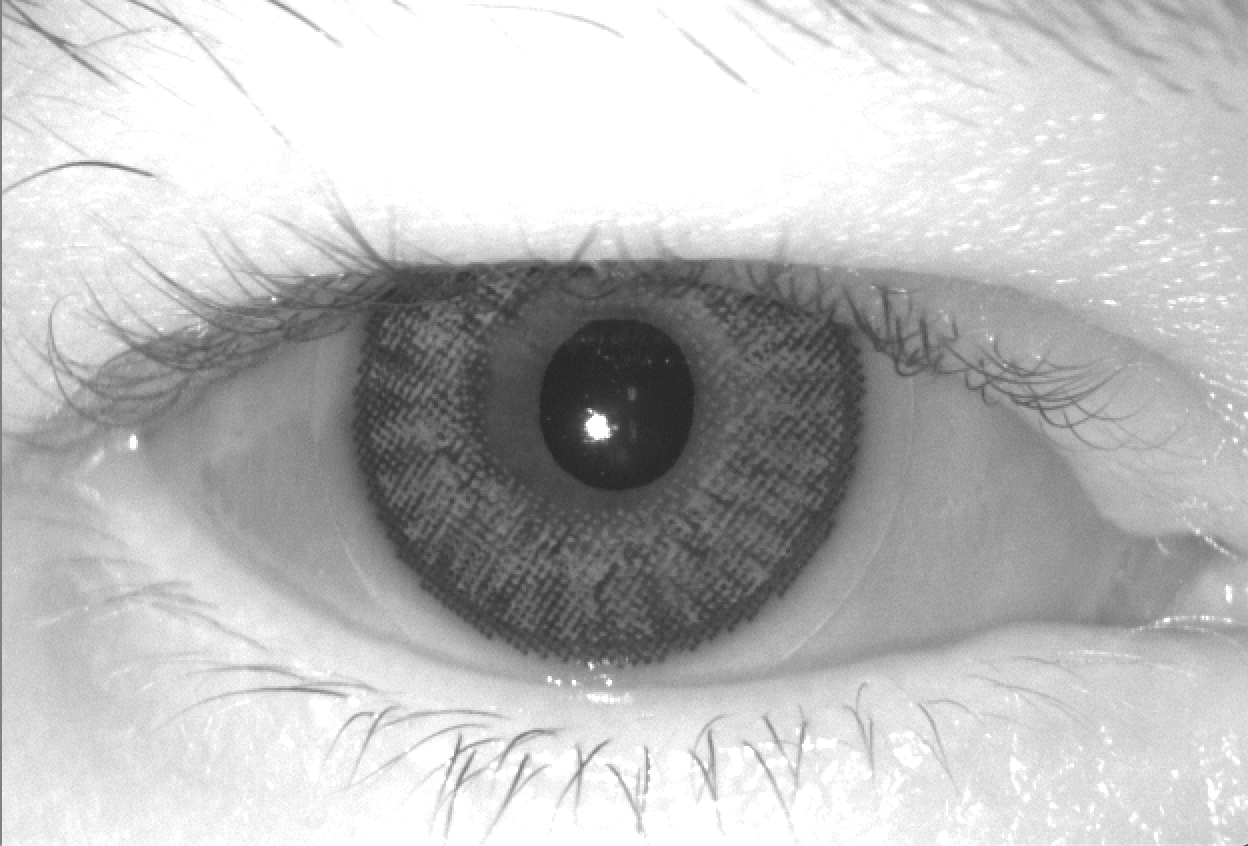}}
    \hfill
    
\caption{Images examples of presentation attack instruments used in LivDet-2020.}
\label{fig:pads}
\end{figure*}

A more recent set of algorithms have been presented at the LivDet-2020 competition \cite{livdet2020}. The best method was proposed by Tapia et al.~\cite{tapia2021iris}. They have presented an approach which have achieved an Average Classification Error Rate (ACER) of 29.78\%. This method also achieved the lowest Bona Fide Classification Error Rate (BPCER) of 0.46\% out of all participant. This paper have shown the relevance of focusing mainly on the bona fide images as a \enquote{first-filter}. However, a broad space for improvement was detected in identifying the PAIs scenarios, especially in cadaver and printed iris images. Figure \ref{fig:pads} shows an example of different PAI's used during training by the winning algorithm in the LivDet-2020 competition. 

In order to achieve better PAD algorithms, it is important to train suchs algorithms with all sort of PAI that could be used to fool biometric systems. The most common Presentation Attack Instruments are created using Contact lenses with a printed pattern or printouts of an iris ~\cite{livdet}. Presenting a print paper with the irises is one of the easiest and cheaper techniques to be used. Printouts can be also produced in many ways. For instance, there is no consensus on whether color or black and white printing is significantly better, or whether matte or glossy paper is better, to make a successful PAI~\cite{OpenSource}.  Building PAI's is not trivial and it usually requires a great effort. Even the creation of PAI using printed eyes can be time demanding in order to get large data sets with different quality papers. That is even more complicated when other methods like those using images from cadavers, contact lenses, plastic contact lenses, or unknown scenarios are used.

The generation of artificial images can be an alternative to create novel PAI's that can be used to train better PAD algorithms. Generative Adversarial Networks (GAN) is a promising techniques that allows to synthesize novel images from a data set.

\subsection{GANs for synthetics images}

The GAN algorithm was first introduced by Ian Goodfellow et al. \cite{survey_gan} and it approaches the problem of unsupervised learning by simultaneously training two deep networks, called Generator $G$ and Discriminator $D$ respectively. These networks compete and cooperate with each other. While the generator creates new instances of the data, the discriminator evaluates them for authenticity. In the course of training, both networks learn to perform their tasks.
To learn a generator distribution $p_{g}$ over data $x$, the generator builds a mapping function from a prior noise distribution $p_z(z)$ to data space as $G(z;\theta_g)$. The discriminator $D(x;\theta_d)$, on the other hand,  outputs a single scalar representing the probability that $x$ came from training data rather than $p_{g}$. During training  parameters of $G$ to minimize  $log(1-D(G(z)))$ while adjusting parameters of $D$ to minimize $log(D(x))$ were simultaneously looked for as if they were following the two-player min-max game with value function $V (G, D)$:


\begin{figure}[H]
\centering
\includegraphics[scale=0.55]{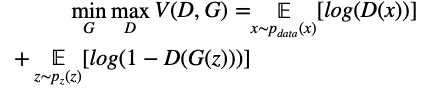}
\end{figure}

There are several variants of The GAN algorithm. Chen et al.~\cite{Chen}, for instance, have proposed a method called Info-GAN. This is an information theoretic extension applied to GAN. This method is able to learn disentangled representations in a completely unsupervised manner. Info-GAN also maximizes mutual information between a small subset of the latent variables and the observation. 

Zhu et al.~\cite{ZhuPIE17} proposed an approach for learning to translate an image from a source domain $X$ to a target domain $Y$ in the absence of paired examples. The goal is to learn a mapping $G : X$ → $Y$ such that the distribution of images from $G(X)$ is indistinguishable from the distribution $Y$ using an adversarial loss. Qualitative results are presented on several tasks where paired training data does not exist, including collection style transfer, object transfiguration, season transfer, photo enhancement. 

Wang et al.~\cite{wang} have presented a method based on pix2pix network \cite{pix2pix} called pix2pixHD. This GAN synthesizing high resolution images from semantic label maps using cGAN. The cGANs have enabled a variety of applications, but the results are often limited with to low resolution and still far from realistic. The author used a generator with $2048\times1024$ visually appealing results with a novel adversarial loss, as well as new multi-scale generator and discriminator architectures.

Yingying et al.~\cite{WGAN} have introduced Wasserstein GANs (WGAN), an alternative to traditional GAN training. In this model, the authors can improve the stability of learning, get rid of problems like mode collapse, and provide meaningful learning curves useful for debugging and hyperparameter searches. Furthermore, they also show that the corresponding optimization problem is sound, and provides extensive theoretical work highlighting the deep connections to other distances between distributions. Afterward, Gulrajani et al. \cite{WANGP} propose a new method Wasserstein GAN-GP (WANGP) to improve the training process in order to get better results and stability the results.

Karras et al. developed the StyleGAN2~\cite{stylegan2}. This GAN is an extension of the progressive growing GAN that is an approach for training generator models capable of synthesizing very large high-quality images via the incremental expansion of both discriminator and generator models from small to large images during the training process. In addition to the incremental growth of the models during training, the StyleGAN2 changes the architecture of the generator significantly. 
The StyleGAN2 generator no longer takes a point from the latent space as input; instead, there are two new sources of randomness used to generate a synthetic image: a standalone mapping network and noise layers. StyleGAN2 introduces the mapping network $f$ to transforms $z$ into this intermediate latent space $w$ using eight fully-connected layers. This intermediate latent space $w$ can be viewed as the new $z$, $(z’)$. Through this network, a 512-D latent space $z$ is transformed into a 512-D intermediate latent space $w$.

Generative adversarial nets have also been extended to conditional models~\cite{cond_Mirza, Chen, ZhuPIE17,kimura20} if both the generator and discriminator are conditioned with extra information $y$. The introduction of external information allows specific representations of generated images to be created. The cGAN is a variant of standard GANs that were introduced to augment GANs with the capability of conditional generation of data samples based on both latent variables (or intrinsic factors) and known auxiliary information (or extrinsic factors). Extrinsic factors could be class information or associated data from other modalities. In other words, cGANs are generative models which can produce data samples $x$, conditioned on both latent variables $z$ and known auxiliary information $y$. 

\subsection{GANs Applied to Iris Images}

GAN algorithms have also been applied to generate synthetics gender labelled iris images. Tapia et al.~\cite{TAPIA2019503}, for instance, proposed E-DCGAN, a conditional GAN algorithm that preserves gender information while generating synthetic images from periocular NIR images. They have shown that standard GAN algorithm are not able to preserve soft biometric features such as gender, while using conditional information allows to achieve a gender labeled synthetic data base. Such synthetic database demonstrated to improve gender classification algorithms when used to augmented the training data set. 

Yadav et al.~ \cite{rasgan} have also proposed a new technique for generating synthetic iris images and demonstrate its potential for Presentation Attack Detection (PAD). The proposed technique improved the loss function to generative capability of a Relativistic Average Standard Generative Adversarial Network (RaSGAN) to synthesize high quality NIR images previously aligned and cropped (iris) to $256 \times 256$ pixels.

One of the difficulties with GAN algorithms, and in particullarly when applied to Iris images or biometrics in general, is how to assess quality and meaningful of the resulting (synthezesed) images.  Only recently, a suite of qualitative and quantitative metrics have been developed to assess the performance of a GAN model based on the quality and diversity of the generated synthetic images \cite{SalimansGZCRC16, HeuselRUNKH17, XU}.  As part of such metrics are: The Inception Score (IS)~\cite{SalimansGZCRC16}, Frechet Inception Distance (FID)~\cite{HeuselRUNKH17,BrockDS19} and Perceptual Path Length (PPL)~\cite{stylegan2}. These metrics allow to compare results from different GAN models. 

FID and IS are both based on feature extraction (the presence or the absence of features). In this work, four state of the art GAN algorithms are used to generate synthetic PAD from a small set of periocular Iris NIR images. The FID metric is used to compare and evaluate quality of the resulting set of images. The resulting PAI, is consider as an unknown attack and used to test the best algorithm from the LiveDet-2020 competition \cite{livdet2020}. 

\section{Proposal: A synthetics Periocular Iris PAI (SPI-PAI)}
\label{sec:proposal}

\begin{figure*}[h]
\centering
\includegraphics[scale=0.50]{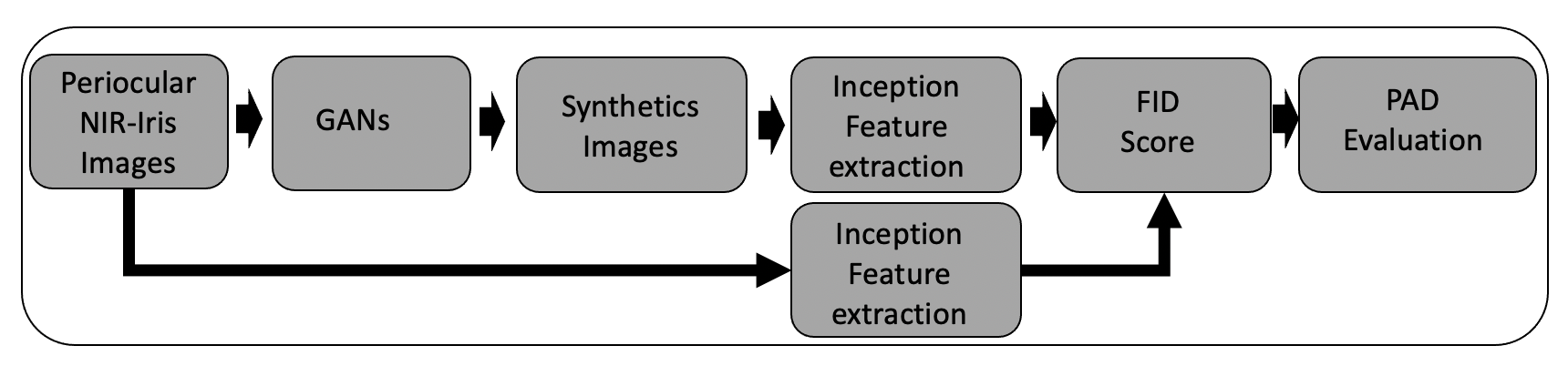}
\caption{Proposal framework to generated a Synthetic PAI using GAN algorithms. The synthetic images are evaluated using the FID score and the best PAI is tested using the state of the art PAD algorithms}
\label{flow_metric}
\end{figure*}

This paper explores the generation of synthetics periocular iris images with four state-of-the-art methods: cGAN, Wasserstein GAN, Wasserstein Gradient Penalty GAN, and StyleGAN-2 (Section \ref{GANS}) using a small dataset of periocular Iris NIR images. In order to compare quality of the generated synthetic images a FID metric is chosen (Section \ref{sec:eval_metrics}). Finally, the best images created were evaluated using a Live Iris image detector algorithm proposed by Tapia et al. at the LivDet-2020 competition \cite{tapia2021iris}. As results, a novel Synthetic Periocular Iris PAI (SPI-PAI) was created from the best synthetic images generated. This new data set will be made available for researcher upon request and it is expected it will contribute to the creation of new PAD systems. A flow chart of the work presented in this paper is shown in Figure \ref{flow_metric}.

\subsection{Database}

For this paper, and for all GAN methods implemented, the GFI-UND database is used \cite{tapia-iriscode}. This database is organized in 3,000 NIR periocular Iris images with a resolution of $640\times480$, captured with an LG-4000 capture device. The database is equally distributed with 1,500 left and 1,500 right NIR iris images. The database is also gender-balanced with a subject-disjoint set of 750 male and 750 female. For training all methods, the input to the GANs were 3,000 images. A probabilistic data-augmentation based on imgaug library \cite{imgaug} with a high occurrence rate $(p: 0.75)$ was used in all experiments.

\subsection{GAN Algorithms benchmark}
\label{GANS}
Unlike other deep learning neural network models that are trained with a loss function until convergence, a GAN generator model is trained using a second model called a discriminator that learns to classify images as real or generated. Both the generator and the discriminator model are trained together to maintain an equilibrium. 
In this work, four state of the art GANs methods are implemented and tested to create synthetics NIR pericocular Iris images, such methods are:
\begin{itemize}
    \item \textit{Conditional GAN} (cGAN).   Conditional GAN was proposed by Mirza et al.~\cite{cond_Mirza} and it considers conditional information. In this case, the conditional information used is the gender. The database contains labels for images in each class of gender (Female and Male). All images from the data set are used as input with its respective labels.
    \item \textit{Wasserstein GAN} (WGAN). This methods was proposed by \cite{WGAN} and despite the cGAN does not consider conditional information. Therefore, the distinction between images from male and female are not made. All images in the training database are used as input to generate the synthetic images.
    \item \textit{Wasserstein Gradient Penalty GAN} (WGAN-GP). This methods is an improvement from previous (Wasserstein GAN) and does not include conditional information, all images are considered from the same class.
    \item \textit{StyleGAN2}. StyleGAN2 was proposed by \cite{stylegan2} and differ from the above techniques mainly in two characteristics. First, this methods used an internal steps that evaluates the quality of the generated images using the FID distance between the generated images and the database used for training (See FID distance in section \ref{sec:eval_metrics}). Seconds, a seed is generated (instead of a set of images) which can be used to synthesize new images from the same class that was used to train the model (with similar mean and variance).
\end{itemize}
A novel PAI is created from each of these techniques. However, In order to achieve a benchmark of the above algorithms and to chose the best one, the FID metric is used. This metric measures similarity of the synthetic images in comparison to the original class \cite{HeuselRUNKH17}. See Figure \ref{flow_metric}.

\subsection{Evaluation Metrics}
\label{sec:eval_metrics}

There is no objective loss function used to train the GAN generator models and no way to assess the progress of the training objectively and the relative or absolute quality of the model from loss alone. In most of the methods, the resulting images looks like the original class of images but specific features are not necessary preserved. This is particularly challenging for periocular images, since the resulting synthesized PAI can look like an eye but soft-biometric information, features or identity information may not be preserved. 

In FID, a pre-trained Inception network is used to extracts features from an intermediate representation from the Inception pre-trained network. The data distribution for these features is modeled using a multivariate Gaussian distribution with mean $u_{x}$ and covariance $\epsilon$. Similarly, a Gaussian distribution is modeled for the original image $G(u_{g}$,$\epsilon_{g})$. In order to compare both Gaussian distribution, the FID metric is applied as it is shown in the following expression:

\begin{equation}
\begin{split}
    FID(x,g)= || u_{x}-u_{g}||^2 + Tr({\epsilon_{x}
+\epsilon{g} -2(\epsilon_{x}\epsilon_{g})^{1/2})})
\end{split}
\end{equation}

Where $Tr$ sums up all the diagonal elements. FID is more robust to noise than the IS metric. FID is also a better measurement for image diversity than other previously reported metrics \cite{BrockDS19}. 

When computing the FID between training and synthetically generated dataset the expected output should be close to zero. As minimum the FID, the closer are the synthesized images to the original used for training. In this work, this metric is used to compared the synthetics images generated using the four state of the art GAN algorithms describes above. When using cGAN, WGAN and WGAN-GP, The FID score is computed between the original set of images used for training the algorithms and the resulting generated images.

\subsection{Unknown Attack}
\label{Fonafide}
Each GAN method used in this paper achieved a set of new synthetic images (Periocular NIR Iris PAI). In order to evaluate such images the Presentation Attack classification USACH/TOC algorithm proposed by \cite{livdet2020} was used. 
This algorithm obtained the first place in the LivDet-2020 competition. This model was trained with 12,000 attack presentations images and it is able to discriminate bona fide images from impostor images. Note that only real images were used as bona fide images to train this model (none synthetics images were used ). The original model reported a Bonafide Presentation Classification Error Rate (BPCER) of 0,46 at the LivDet-2020 competition. This algorithm will be used to test if the synthetics images generated (Periocular NIR Iris PAI) are classified as bona fide or impostor images (unknown Attack).

\section{Experiments and Results}
\label{sec:exp_results}
\subsection{Synthetics Image generation}
As mentioned before, this experiment focuses on obtaining high-quality synthetics images using four states of the art GAN-based algorithms. All the experiments were performed with Intel Xeon E5 and GPUNvidia Tesla V100 16GB.
As follows, four experiments (one for each GAN method:cGAN, WGAN, WGAN-GP, and StyleGAN2) are performed:

\subsubsection{Experiments 1 - cGAN:}
The cGAN methods was implemented and trained using two sizes of images, $80\times160$ and $320\times240$ pixels. The FID measure was computed for all resulting datasets of synthetic images. Table \ref{table-cgan} shows results (FID obtained ) for several experiments using a resolution of  $80\times160$ and $320\times240$ respectively.

In most of the experiments, synthetic images generated using a resolution of $80\times160$ present an inferior similarity to the original dataset than those of size $320\times240$ pixels. The FID score is higher for most of such experiments. The best FID score, was achieved when images of $320\times240$ pixels are used and it reaches a value of $159,12$. This result was obtained using 500 epochs and 9,000 synthetic images. 
\begin{table}[H]
    \centering
    \caption{\label{table-cgan} Summary of the FID scores for images synthesized using cGAN and image size of $80\times60$ pixels. KIMGS refers to the number of synthetic images generated and used for computing the FID Score}
    \begin{tabular}{|c|c|c|c|}
    \hline
        \textbf{Epochs} &  \textbf{KIMGS}  & \textbf{$80\times60$ pixels}   & \textbf{$320\times240$ pixels} 
        \\ \hline
        100   & 3,0    & 317,41  &  343,12 \\ \hline
        250   & 6,0.   & 240,43  &  241,22  \\ \hline
        500   & 9,0    & 246,21  &  159,12  \\ \hline
        1,000 & 12,0   & 312,83  &  242,62   \\ \hline
        1,500 & 15,0   & 412,33  &  342,80  \\ \hline
    \end{tabular}
\end{table}

\subsubsection{Experiment-2 Wasserstein GAN}

The W-GAN method was implemented and trained using images of $320\times240$ pixels. 
Several experiments using different epochs and different number of synthetic images were tested.

Table \ref{table-wgan} shows a summary of the FID score reached for these experiments. The best results (minimum FID score) for the synthetic images generated was 97.66\% and it was reached with 500 epochs, 9,000 synthetic images and the followings parameters: Learning rate of $1e-5$, Optimizer: RMsprop and Batch size of $60$. 
At first glance, under similar experiment conditions this methods generated a substantial improvement with respect to the previous one in terms of FID score.

\begin{table}[H]
    \centering
    \caption{\label{table-wgan}Summary of the FID scores for WGAN for image size of $320\times240$ pixels.}
    \begin{tabular}{|c|c|c|}
    \hline
        \textbf{Epochs} &  \textbf{KIMGS}  & \textbf{FID Score}  \\ \hline
        100 & 3.0 & 227,66      \\ \hline
        250 & 6.0 & 127,66      \\ \hline
        500 & 9.0 & 97,66       \\ \hline
        1,000 & 12.0 & 125,14     \\ \hline
        1.500 & 15.0 & 250,90   \\ \hline
    \end{tabular}
\end{table}

\subsubsection{Experiment 3 - Wasserstein Gradient Penalty GAN}

The WGAN-GP method was implemented and trained using 3,000 input images of $320\times240$ pixels.

Table \ref{table-wgangp} shows a summary of the FID score reached for the  synthetic images generated using this method. Several experiments using different epochs and different number of synthetic images were tested. The best results (minimun FID score) were reached with 12,000 synthetic images and the followings parameters: Learning rate of $1e-4$, Epochs: $1,000$, Optimizer: Adam, and Batch size of $60$. 

\begin{table}[H]
    \centering
    \caption{\label{table-wgangp} Summary of the FID scores for synthetic images generated using  WGAN-GP  (image size of $320\times240$ pixels).}
    \begin{tabular}{|c|c|c|}
    \hline
        \textbf{Epochs} &  \textbf{KIMGS}  & \textbf{FID Score}  \\ \hline
        100   & 3,0     & 196,81   \\ \hline
        250   & 6,0     & 184,19   \\ \hline
        500   & 9,0     & 134,22   \\ \hline
        1,000   & 12,0  & 92,12   \\ \hline
        1,500 & 15,0    & 209,32   \\ \hline
    \end{tabular}
   
\end{table}

\subsubsection{Experiment 4 - \textit{StyleGAN2}}

Several hyper-parameters were explored with StyleGAN2 in order to improve the quality of the images and reduce as much as possible the FID score in between the generated images and the original data set. Despite previous methods, StyleGAN2 shown to be very sensitive to changes in hyper-parameters, specially learning rate.

\begin{figure}[H]
	\centering
	\includegraphics[scale=0.30]{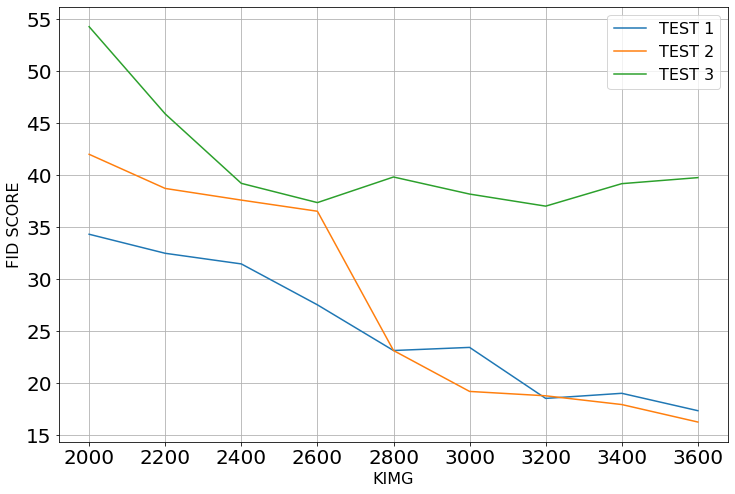}
	\caption{\label{fig:3fid} Summary of the exploration of the FID with different learning rates.}
\end{figure}

Figure \ref{fig:3fid} shows the results of the StyleGAN2 whit different learning rates: Test one (blue) represents a learning rate of $2.5e-3$, test two (Orange) represents a learning rate of $1.0e-4$ and test three (Green) represents a learning rate of $1.0e-2$. 
The Y-axis in the figure shows the FID score, and X-axis shows the number of thousand images (KIMG) generated for the network during training.

The best results for StyleGAN2 were reached with the followings parameters: Learning rate of $2.5e-3$, Epochs: 1,200, Optimizer: Adam and Batch size of $60$. The images were used with a resolution of $320\times240$ pixels.

Table \ref{table-stylegan2} shows a summary of the FID score reached for several experiments using StyleGAN2 where the number of synthetic images is increased (along with the number of epochs). The FID score was calculated every 200 KIMGS. It can be appreciate that the FID score decreases when number of generated images (KIMGS) are increased.  The best FID score is reached with 3,600 KIMGS which correspond to 3,6 millions of generated synthetic images.


\subsection{Algorithms Benchmark}

\begin{figure*}[h]
\centering
\begin{tabular}{cccc}
   (a) cGAN  & (b) W-GAN &  (c) WGAN-GP & (d) StyleGAN2\\

\includegraphics[scale=0.75]{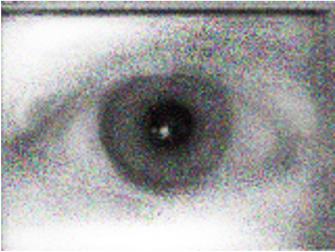} &
\includegraphics[scale=0.36]{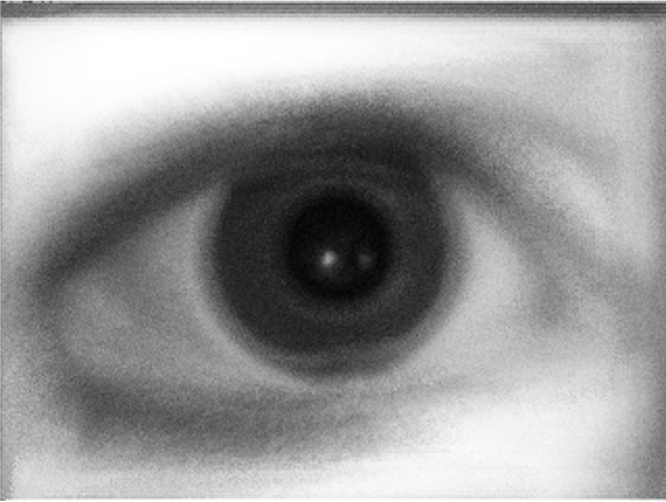}&
\includegraphics[scale=0.36]{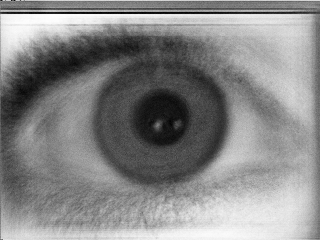}&
\includegraphics[scale=0.34]{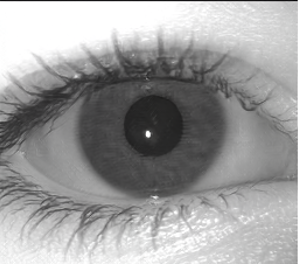}
\\
\includegraphics[scale=0.75]{images/cgan-iris/mejor-2-cgan.png} &
\includegraphics[scale=0.36]{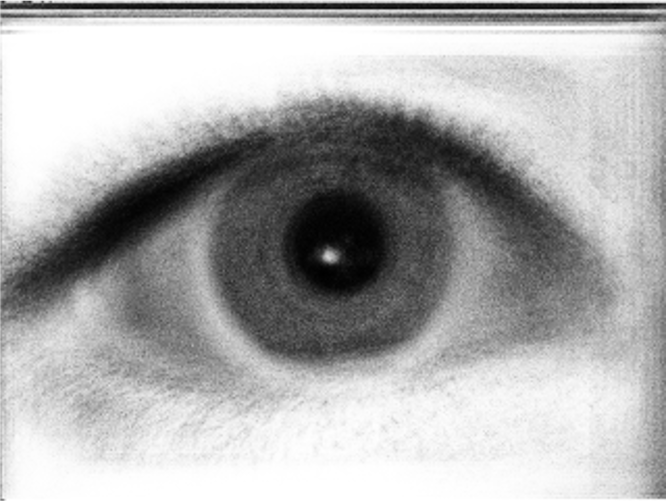}&
\includegraphics[scale=0.36]{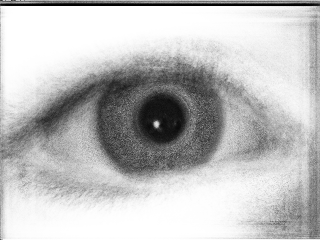}&
\includegraphics[scale=0.34]{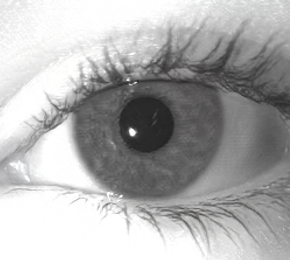}
\end{tabular} \\
\caption{Example of synthetics images generated using experiments 1 to 4. Left to right: cGAN, W-GAN, WGAN-GP and StyleGAN2.}
\label{cgan-example}
\end{figure*}

As it was shown in previous experiments, StyleGAN2 achieved the minimum FID score from all four GAN methods. Figure \ref{cgan-example} shows of the images synthesized using the four GAN methods. As it can be appreciated, the method with higher FID score (cGAN) achieved low quality images (first column in Figure \ref{cgan-example}). In such images, it is not possible to distinguish the pupil of the iris or other components of the eye in general. The image is quite noisy. Wasserstein GAN methods (columns 2 and 3 in Figure \ref{cgan-example}) show important improvements. In both cases the features of the eye are better defined. However, the eyelashes are blurry and can not be distinguished. The StyleGAN2 algorithm, on the other hand, achieved the best performance from all  GAN methods. The minimum FID score reached was 16.29. The resulting images of this method are clearly similar to the images they where generated from (GFI-UND-database \cite{tapia-iriscode}). In this case, not just the iris, pupil and sclera are well defined but also the eyelashes. 

\begin{table}[H]
\centering
\caption{\label{table-stylegan2}Summary of the FID scores for StyleGAN2 for image size of $320\times240$ pixels. The lower FID score was 16.29 with 3,600 KIMGS.}
    \begin{tabular}{|c|c|}
    \hline
        \textbf{KIMGS}  & \textbf{FID Score}  \\ \hline
         200  & 352.34 \\ \hline
         400  & 337.78 \\ \hline
         600  & 281.34 \\ \hline
         800  & 237.16 \\ \hline
         1,000  & 138.18 \\ \hline
         1,200  & 98.81 \\ \hline
         1,400  & 78.91 \\ \hline
         1,600  & 73.43 \\ \hline
         1,800  & 61.37 \\ \hline
         2,000  & 42.01 \\ \hline
         2,200  & 38.73 \\ \hline
         2,400  & 37.61 \\ \hline
         2,600  & 36.54 \\ \hline
         2,800  & 23.16 \\ \hline
         3,000  & 19.23 \\ \hline
         3,200  & 18.81 \\ \hline
         3,400  & 17.97 \\ \hline
        \textbf{3.600}  & \textbf{16.29} \\ \hline
         3,800  & 27.91 \\ \hline
    \end{tabular}
\end{table}

\subsection{Unknow Atack - PAD testing}

In this section the best synthetic PAI generated (using StyleGAN2) is evaluated using a PAD of two classes, Bona fide and Presentation Attack \cite{livdet2020}. This PAI corresponds to 3,000 synthetic images generated from a single seed created using the the best StyleGAN2 model. Figure \ref{collage_stylegan} shows examples of the images of this synthetics Periocular Iris PAI (SPI-PAI). The first image (left corner) is the primary reference. The FID score reached for this data set was 16.29 (As reported in Table \ref{table-stylegan2}).

Two experiments are performed: First, the SPI-PAI is presented to the PAD as unknown attack. Surprisingly, all synthetics images were classified as bona fide. Therefore, the algorithm was not able to detect this images as synthetic and it is completely fooled by the generated SPI-PAI.

In the second experiment, the PAD  was tested using the synthetic PAI generated (SPI-PAI ) and an equal set (in quantity) of impostor images. In this case the Bona fide Presentation Classification Error Rate(BPCER) was 0,0\%,  Attack Presentation Classification Error Rate (APCER) of 0.3244, and Attack Classification Error Rare (ACER) of 0.1603 according to the requirements of ISO/IEC 30107-3. 

A similar experiment was performed for each PAI created with all four algorithms, in each case, the Detection Equal Trade-off Curve (DET) was computed and it is reported in Table \ref{tab:EER}.

\begin{table}[H]
\caption{}
\centering
\label{tab:EER}
\begin{tabular}{|l|l|l|l|l|}
\hline
                          & cGAN                    & W-GAN                   & WGAN-GP                 & StyleGAN2                 \\ \hline
\multicolumn{1}{|c|}{D-EER \%} & \multicolumn{1}{c|}{41,1} & \multicolumn{1}{c|}{35,02} & \multicolumn{1}{c|}{25,01} & \multicolumn{1}{c|}{10,01\%} \\ \hline
\end{tabular}
\end{table}

In agreement with results shown for the FID score computed for each method, the StyleGAN2 algorithm achieved better synthetic periocular NIR images and can easily fool the state of the art PAD algorithm. 

\begin{figure}[]
	\centering
	\includegraphics[scale=0.32]{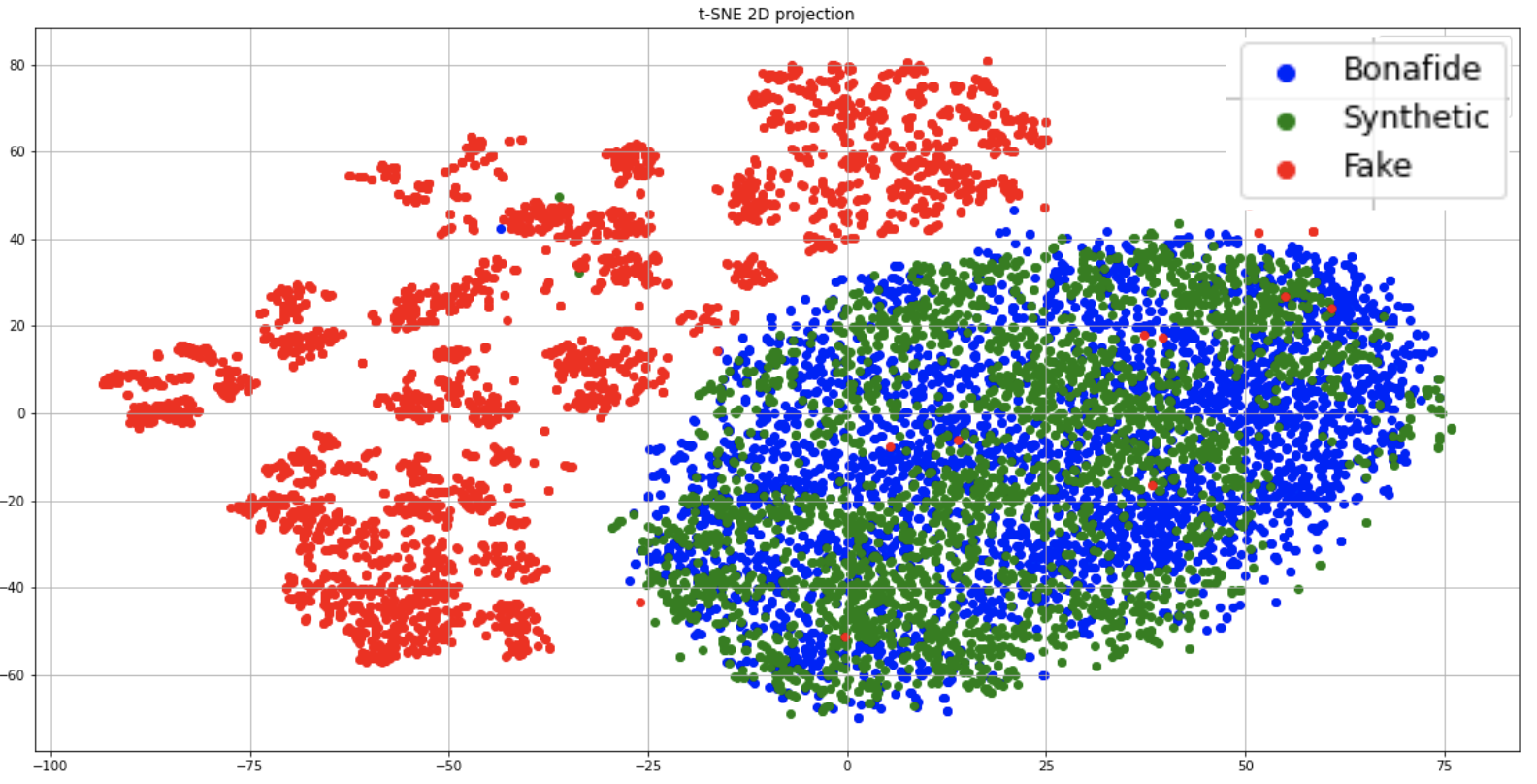}
	\caption{\label{collage_stylegan} 2D Projection using t-SNE visualization of bona fide, synthetic and Fake images.}
	\label{fig:tsne}
\end{figure}

Figure \ref{fig:tsne} shows a t-distributed Stochastic Neighbor Embedding (t-SNE) map. This method shows non-linear connections in the data.  The 2D maps shows a representation of bona fide (Blue), synthetics (Green) and fake images (Red) dots. The bona fide  and synthetics information are highly embedded in the feature space as the figure shows. The t-SNE algorithm reveals a behavior according to the similarity measures of the images (bona fide versus synthetic) which results in a  FID score of 16,29.

The fake images, on the other hand, are clustered in a completely separate space which is in concordance with the higher FID score measured with respect to the synthetic images (FID = 120.73). 

According to ISO/IEC 29794-6 standard, the sharpness measures the degree of focus present in the image. Sharpness is measured as a function of the power spectrum after filtering with a Laplacian of Gaussian operator. The mean sharpness of synthetics images was 30.63 and the mean sharpness synthetics images was 29.362. The similarly of these metrics demonstrated that the synthetic images well preserve the quality according to the ISO/IEC 29794-6 standard. Therefore, such images can be used for biometric applications.

\section{Conclusion}
\label{sec:conclusion}
It is not trivial to create synthetic images from periocular NIR images. The resulting images depend on the generation method and the resolution used. The resulting images can look like periocular iris images, but the quality may not be enough to be used to train presentation attacks algorithms (i.e: blur, noise, artifacts). In this paper, four state of the art methods for synthetic images generation were tested (cGAN, WGAN,WGAN-GP and StyleGAN2). The resulting images were evaluated using the FID score in order to assess how close to the input class (original images) they are. Results shows that StyleGAN2 not only minimize better the FID distance but also achieve better quality images (i.e: well defined edges, less blur and noise). Additionally, the images were used as Unknown Attack to test the state of the art algorithm (winner of the LivDet-2020 competition). Surprisingly, this algorithm was not able to discriminate the images created as synthetic, classifying all of them as bona fide images (BPCER of 0,0\%). 

\begin{figure*}[]
	\centering
	\includegraphics[scale=0.65]{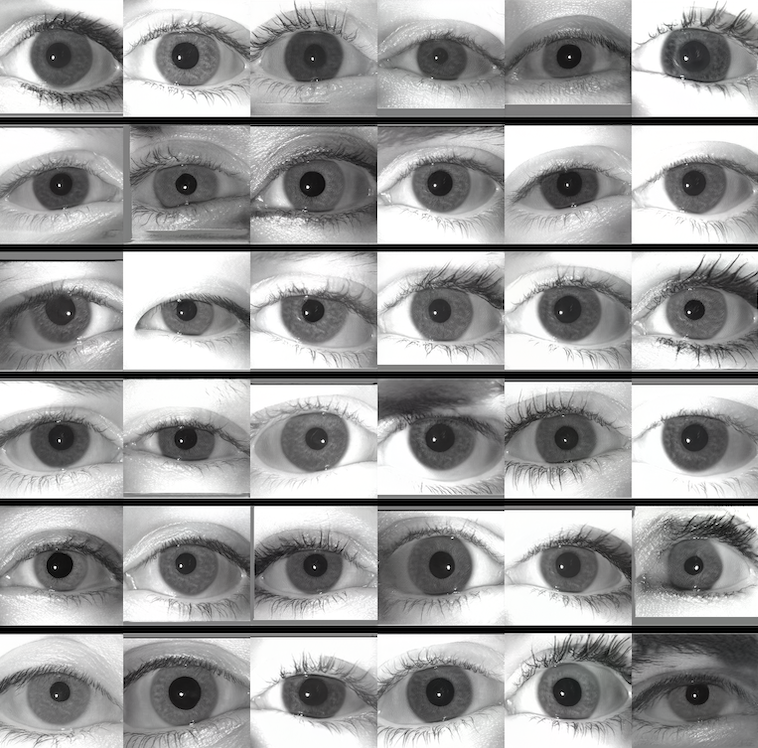}
	\caption{\label{collage_stylegan} Example of 36 synthetics iris images using the best model with StyleGAN2.}
	\label{fig:log_reg}
\end{figure*}

\section{Future Work}
\label{sec:future}
As future work we expect to be able to increase the size of the generated images (mainly dependent on GPU capacity and time) and to explore the creation of new set of synthetic images that can consider novel scenarios such the use of contact lenses and cadaver images among others. It is also important to consider preservation of biometric identity in such images and to explore new metric for comparing and assessing quality of image and performance of the algorithms. This novel PAI has shown to be able to fool PAD systems, therefore, we expect to work in novel algorithms that include such scenario.

\section*{Acknowledgment}
This research work has been partially funded by the German Federal Ministry of Education and Research and the Hessian Ministry of Higher Education, Research, Science and the Arts within their joint support of the National Research Center for Applied Cybersecurity ATHENE and TOC Biometrics company.

\ifCLASSOPTIONcaptionsoff
  \newpage
\fi

\bibliographystyle{IEEEtran}
\bibliography{ReferencesGAN.bib}
\newpage

\begin{IEEEbiography}[{\includegraphics[width=1in,height=1.25in,clip,keepaspectratio]{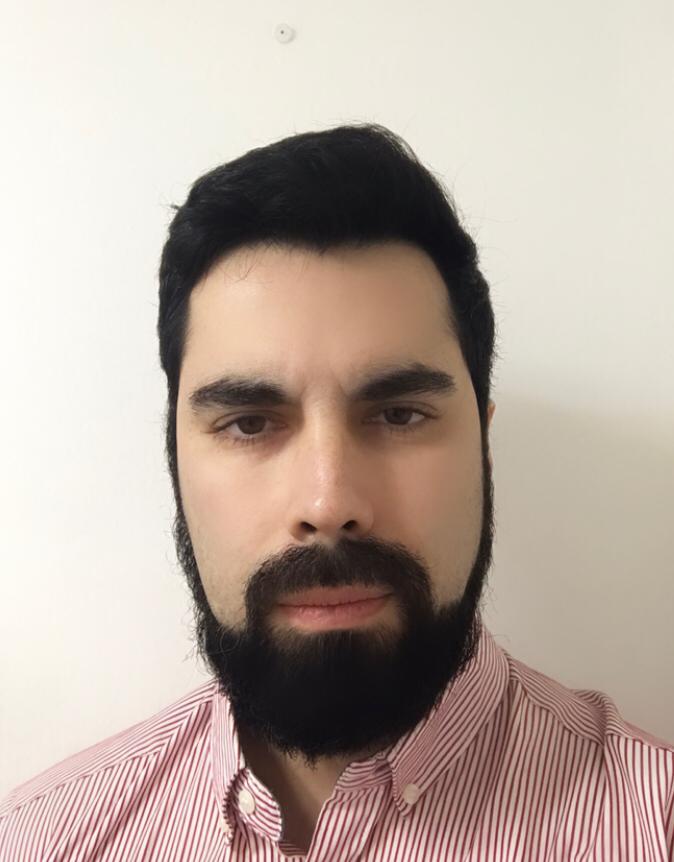}}]{Jose Maureira} received a B.S. in Computer Engineering from Universidad Santiago de Chile, Departmento de informatica, Chile in 2021. His main interests include computer vision, pattern recognition and Deep learning applied to semantic synthetics images.
\end{IEEEbiography}

\begin{IEEEbiography}[{\includegraphics[width=1in,height=1.25in,clip,keepaspectratio]{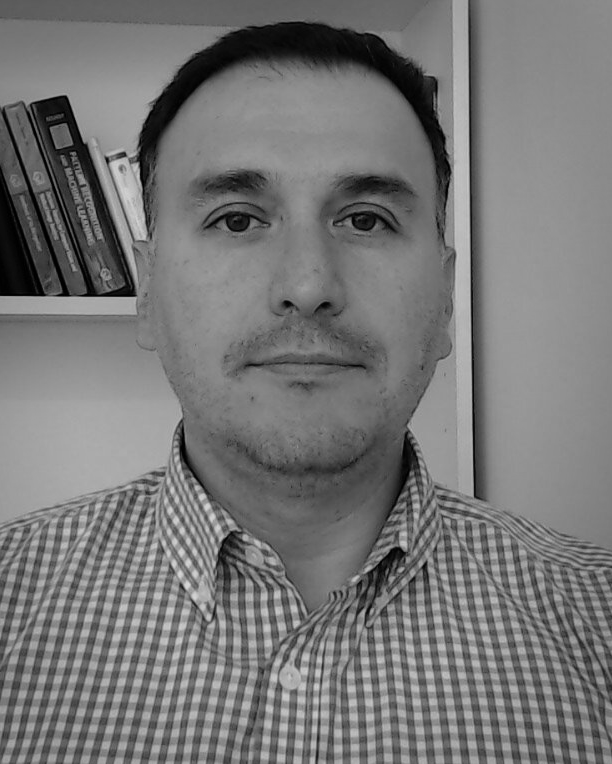}}]{Juan Tapia} received a P.E. degree in Electronics Engineering from Universidad Mayor in 2004, a M.S. in Electrical Engineering from Universidad de Chile in 2012, and a Ph.D. from the Department of Electrical Engineering, Universidad de Chile in 2016. In addition, he spent one year of internship at University of Notre Dame. In 2016, he received the award for best Ph.D. thesis. From 2016 to 2017, he was an Assistant Professor at Universidad Andres Bello. From 2018 to 2020, he was the R\&D Director for the area of Electricity and Electronics at Universidad Tecnologica de Chile - INACAP. He is currently a Senior Researchet at Hoschule Darmstadt(HDA), and R\&D Director of TOC Biometrics. His main research interests include pattern recognition and deep learning applied to iris biometrics, morphing, feature fusion, and feature selection. 
\end{IEEEbiography}

\begin{IEEEbiography}[{\includegraphics[width=1in,height=1.25in,clip,keepaspectratio]{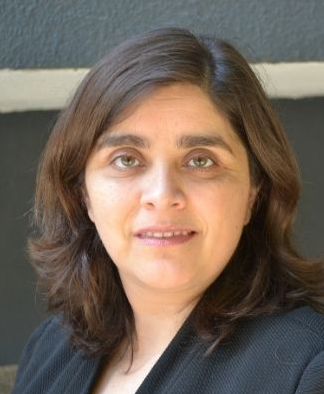}}]{Claudia Arellano} received a Civil Electrical and In- dustrial degree from Pontificia Universidad Catolica de Chile (2003). She also hold a Master of sciences from the same university (2005). In 2014 she was awarded with a Ph.D. from the Trinity College Dublin in Ireland. Her research interests include patterns recognition, shape detection and robust statistics. She is currently an assistant professor at the Business School of Universidad Adolfo Ibañez.
\end{IEEEbiography}

\begin{IEEEbiography}[{\includegraphics[width=1in,height=1.25in,clip,keepaspectratio]{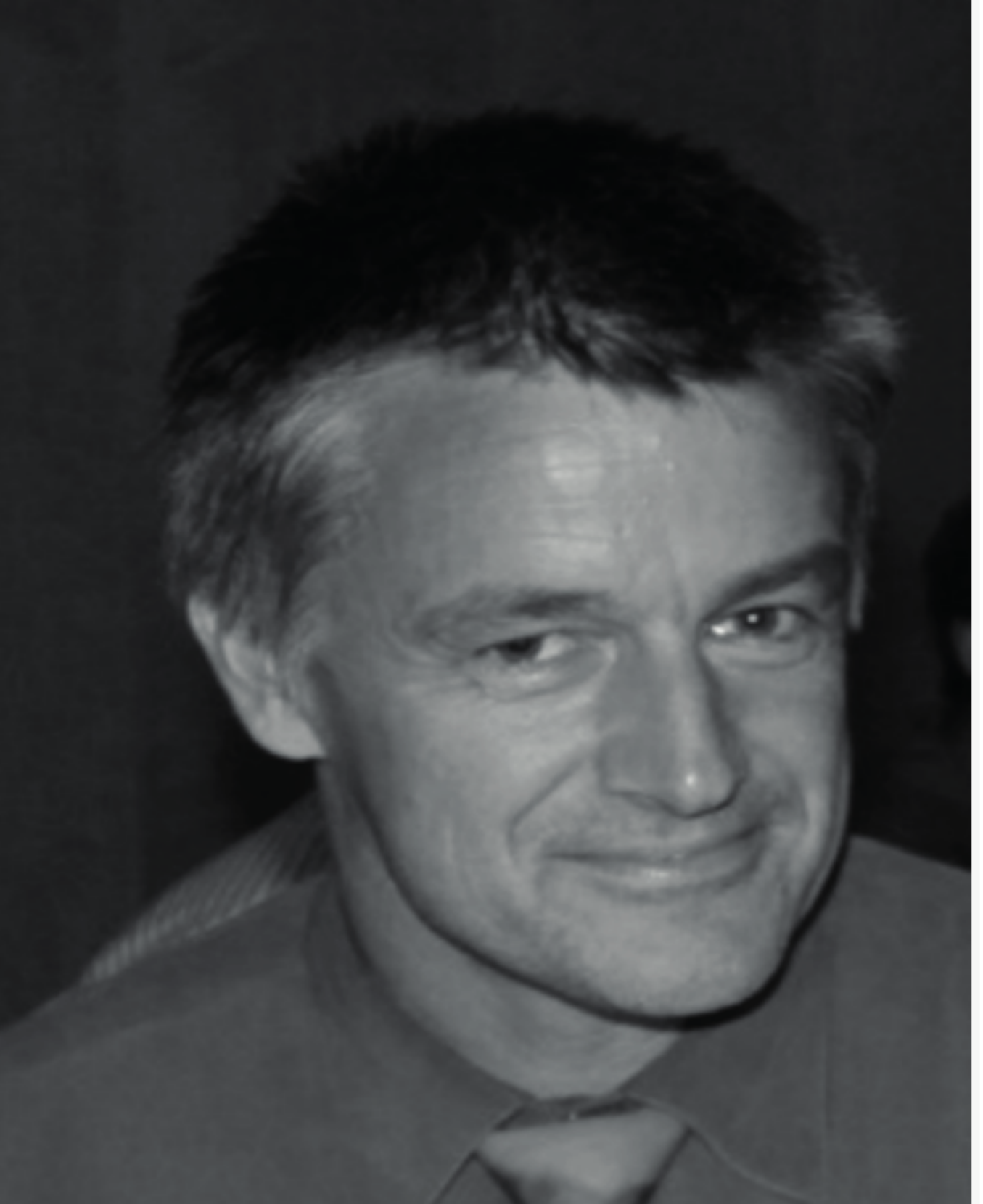}}]{Christoph Busch} is member of the Department of Information Security and Communication Technology (IIK) at the Norwegian University of Science and Technology (NTNU), Norway. He holds a joint appointment with the computer science faculty at Hochschule Darmstadt (HDA), Germany. Further he lectures the course Biometric Systems at Denmark’s DTU since 2007. On behalf of the German BSI he has been the coordinator for the project series BioIS, BioFace, BioFinger, BioKeyS Pilot-DB, KBEinweg and NFIQ2.0. In the European research program he was initiator of the Integrated Project 3D-Face, FIDELITY and iMARS. Further he was/is partner in the projects TURBINE, BEST Network, ORIGINS, INGRESS, PIDaaS, SOTAMD, RESPECT and TReSPAsS. He is also principal investigator in the German National Research Center for Applied Cybersecurity (ATHENE). Moreover Christoph Busch is co-founder and member of board of the European Association for Biometrics (www.eab.org) that was established in 2011 and assembles in the meantime more than 200 institutional members. Christoph co-authored more than 500 technical papers and has been a speaker at international conferences. He is member of the editorial board of the IET journal.
\end{IEEEbiography}

\end{document}